\documentclass[10pt,twocolumn,letterpaper]{article}

\usepackage{cvpr}               

\usepackage{graphicx}
\usepackage{amsmath}
\usepackage{amssymb}
\usepackage{booktabs}
\usepackage{multirow}
\usepackage{xcolor}
\usepackage[accsupp]{axessibility}
\usepackage{subfiles}

\usepackage[pagebackref,breaklinks,colorlinks]{hyperref}

\usepackage[capitalize]{cleveref}
\crefname{section}{Sec.}{Secs.}
\Crefname{section}{Section}{Sections}
\Crefname{table}{Table}{Tables}
\crefname{table}{Tab.}{Tabs.}

\usepackage[normalem]{ulem}

\def\cvprPaperID{1141} 
\def\confName{CVPR}
\def\confYear{2023}

\begin{document}

\title{Egocentric Audio-Visual Object Localization}


\author{
Chao Huang{$^1$}, Yapeng Tian{$^1$}, Anurag Kumar{$^2$}, Chenliang Xu{$^1$} \\
{$^1$}University of Rochester, {$^2$}Meta Reality Labs Research\\
{\tt\small \{chaohuang, yapengtian, chenliang.xu\}@rochester.edu, anuragkr90@meta.com}}


\maketitle

\begin{abstract}
    
Humans naturally perceive surrounding scenes by unifying sound and sight from a first-person view.
Likewise, machines are advanced to approach human intelligence by learning with multisensory inputs from an egocentric perspective.
In this paper, we explore the challenging egocentric audio-visual object localization task and observe that 1) egomotion commonly exists in first-person recordings, even within a short duration; 2) The out-of-view sound components can be created when wearers shift their attention. 
To address the first problem, we propose a geometry-aware temporal aggregation module that handles the egomotion explicitly.
The effect of egomotion is mitigated by estimating the temporal geometry transformation and exploiting it to update visual representations.
Moreover, we propose a cascaded feature enhancement module to overcome the second issue. It improves cross-modal localization robustness by disentangling visually-indicated audio representation.
During training, we take advantage of the naturally occurring audio-visual temporal synchronization as the ``free'' self-supervision to avoid costly labeling.
We also annotate and create the \textit{Epic Sounding Object} dataset for evaluation purposes. 
Extensive experiments show that our method achieves state-of-the-art localization performance in egocentric videos and can be generalized to diverse audio-visual scenes. 
Code is available at \url{https://github.com/WikiChao/Ego-AV-Loc}.

\end{abstract}

\section{Introduction}
\label{sec:intro}

\begin{figure}[!t]
    \centering
    \includegraphics[width=0.47\textwidth]{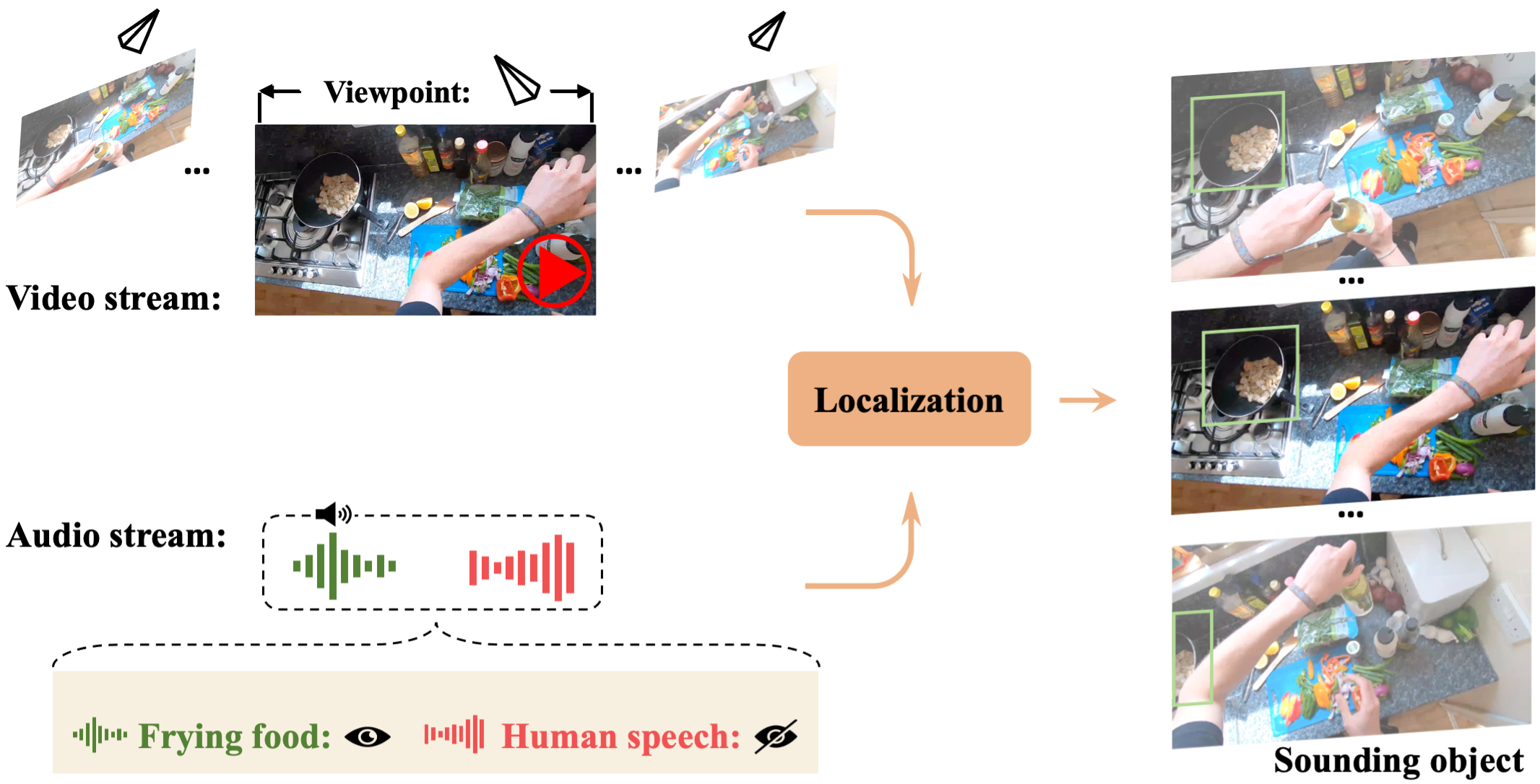}
    \caption{\textbf{Sounding object localization in egocentric videos.} 
    Due to the wearer's egomotion, the viewpoint changes continuously across time.
    Consequently, audio-visual relations are dynamically changing in egocentric videos.  
    Our approach tackles challenges in the egocentric audio-visual sounding object task and learns audio-visual associations from first-person videos.
    } 
    \label{fig:motivation}
    \vspace{-5mm}
\end{figure}


%
%
The emergence of wearable devices has drawn the attention of the research community to egocentric videos, the significance of which can be seen from egocentric research in a variety of applications such as robotics~\cite{martin2021jrdb,kim2019eyes,kawamura2002toward}, augmented/virtual reality~\cite{jones2008effects,poupyrev1998egocentric,swan2007egocentric}, and healthcare~\cite{serino2015detecting,morganti2013allo}.
%
In recent years, the computer vision community has made substantial efforts to build benchmarks~\cite{damen2018scaling,damen2020rescaling,northcutt2020egocom,donley2021easycom,sigurdsson2018charades,li2018eye}, establish new tasks~\cite{fathi2011learning,li2013pixel,li2013learning,park2016egocentric,lee2012discovering}, and develop frameworks~\cite{li2021ego,kazakos2019epic,munro2020multi,wang2021interactive} for egocentric video understanding.

%
%
%

While existing works achieve promising results in the egocentric domain, it still remains an interesting but challenging topic to perform fine-grained egocentric video understanding.
For instance, understanding which object is emitting sound in a first-person recording is difficult for machines. As shown in Fig.~\ref{fig:motivation}, the wearer moves his/her head to put down the bottle. The frying pot which emits sound subsequently suffers deformation and occlusion due to the wearer's egomotion. 
Human speech outside the wearer's view also affects the machine's understanding of the current scene.
This example reveals two significant challenges for designing powerful and robust egocentric video understanding systems:
First, people with wearable devices usually record videos in naturalistic surroundings, where a variety of illumination conditions, object appearance, and motion patterns are shown. 
The dynamic visual variations introduce difficulties in accurate visual perception.
Second, egocentric scenes are often perceived within a limited field of view (FoV). The common body and head movements cause frequent view changes (see Fig.~\ref{fig:motivation}), which brings object deformation and creates dynamic out-of-view content.

Although a visual-only system may struggle to fully decode the surrounding information and perceive scenes in egocentric videos, audio provides stable and persistent signals associated with the depicted events. 
Instead of purely visual perception, numerous psychological and cognitive studies~\cite{shams2008benefits,jacobs2019can,bulkin2006seeing,spence2003multisensory} show that integration of auditory and visual signals is significant in human perception.
%
%
Audio, as an essential but less focused modality, often provides synchronized and complementary information with the video stream.
In contrast to the variability of first-person visual footage, sound describes the underlying scenes consistently.
These natural characteristics make audio another indispensable ingredient for egocentric video understanding.
%

%
To effectively leverage audio and visual information in egocentric videos, a pivotal problem is to analyze the fine-grained audio-visual association, 
specifically identifying which objects are emitting sounds in the scene.
%
In this paper, we explore a novel egocentric audio-visual object localization task, which aims to associate audio with dynamic visual scenes and localize sounding objects in egocentric videos.
Given the dynamic nature of egocentric videos, it is exceedingly challenging to link visual content from various viewpoints with audio captured from the entire space.
%
Hence, we develop a new framework to model the distinct characteristics of egocentric videos by integrating audio.
In the framework, we propose a geometry-aware temporal module to handle egomotion explicitly. 
Our approach mitigates the impact of egomotion by performing geometric transformations in the embedding space and aligning visual features from different frames. 
%
%
We further use the aligned features to leverage temporal contexts across frames to learn discriminative cues for localization.  
Additionally, we introduce a cascaded feature enhancement module to handle out-of-view sounds.
The module helps mitigate audio noises and improves cross-modal localization robustness.

Due to the dynamic nature of egocentric videos, it is hard and costly to label sounding objects for supervised training.
%
%
To avoid tedious labeling, we formulate this task in a self-supervised manner, and our framework is trained with audio-visual temporal synchronization. 
Since there are no publicly available egocentric sounding object localization datasets, we annotate an \textit{Epic Sounding} dataset to facilitate research in this field. 
Experimental results demonstrate that modeling egomotion and mitigating out-of-view sound can improve egocentric audio-visual localization performance.
%

%


In summary, our contributions are: 
(1) the first systematical study on egocentric audio-visual sounding object localization;
(2) an effective geometry-aware temporal aggregation approach to deal with unique egomotion;
(3) a novel cascaded feature enhancement module to progressively inject localization cues; and
(4) an \textit{Epic Sounding Object} dataset with sounding object annotations to benchmark the localization performance in egocentric videos.


\begin{figure*}[t]
    \centering
    \includegraphics[width=0.8\textwidth]{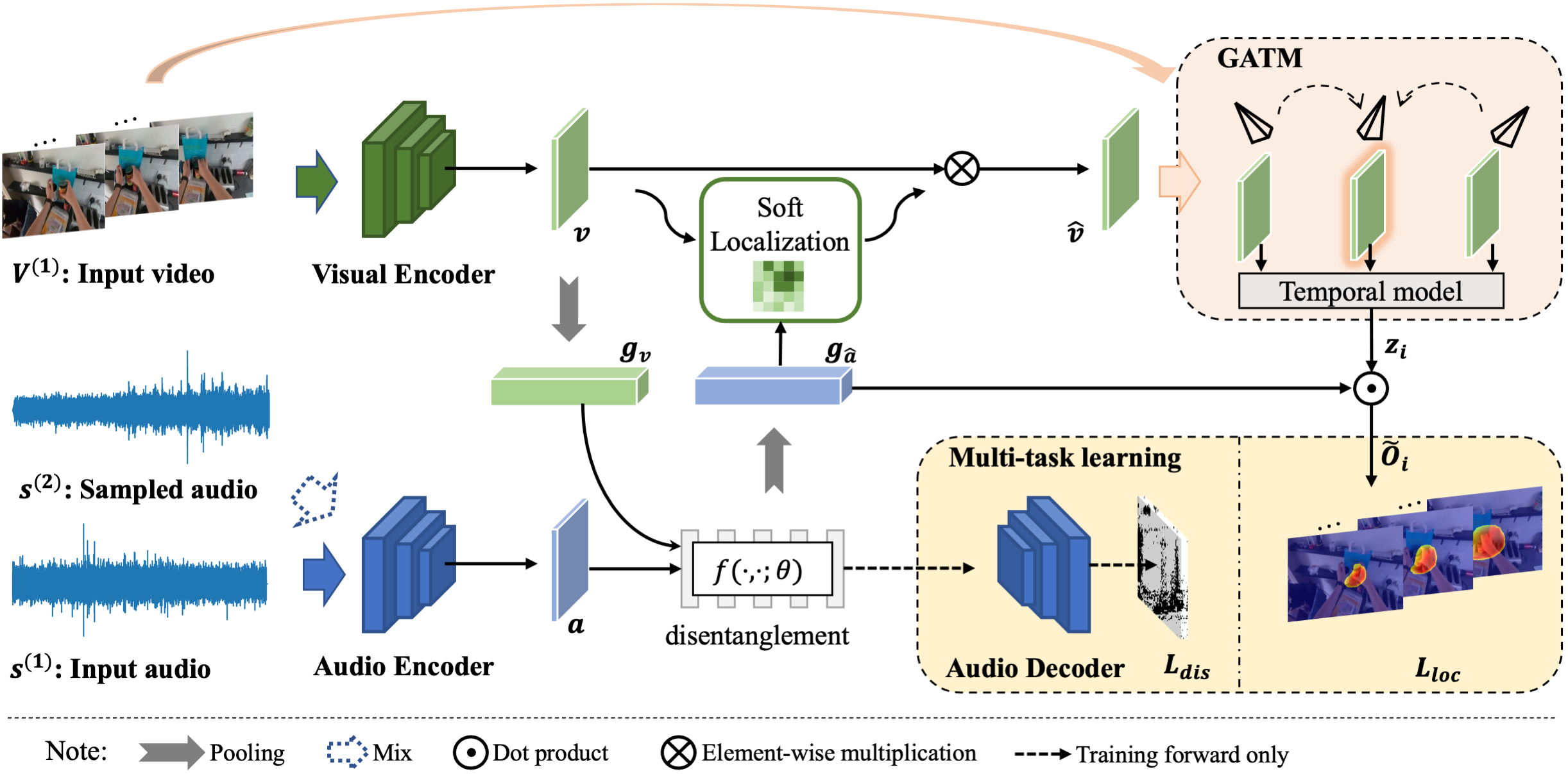}
    \caption{An overview of our egocentric audio-visual object localization framework. In the beginning, our model extracts deep features from the video and audio streams. Then, the audio and visual features are fed into the cascaded feature enhancement module to inject localization cues for both branches. Such a module is additionally trained with ``mix-and-separation" strategy. Next, our geometric-aware temporal modeling block leverages the relative geometric information between visual frames and performs temporal context aggregation to get the final visual features for localization.}
    \label{fig:framework}
    \vspace{-4mm}
\end{figure*}

\section{Related Work}
\label{sec:background}


\noindent \textbf{Audio-visual learning in third-person view videos.}
Taking the natural audio-visual synchronization in videos, a large number of studies in the past few years have proposed to jointly learn from both auditory and visual modalities. 
%
We have seen a spectrum of new audio-visual problems and applications, including visually guided sound source separation~\cite{ephrat2018looking,zhao2018sound,tian2021cyclic,gao2019co,gan2020music,gao2018learning,gao20192,zhou2020sep, rouditchenko2019self}, audio-visual representation learning~\cite{arandjelovic2017look,aytar2016soundnet,hu2019deep,korbar2018cooperative,owens2018audio,owens2016ambient,afouras2020self}, audio-visual event localization~\cite{tian2018audio,tian2019audio,wu2019dual,lin2019dual}, audio-visual video parsing~\cite{tian2020unified,wu2021exploring}, and sounding object visual localization~\cite{senocak2018learning,ots2018,chen2021localizing,qian2020multiple,li2021space,hu2022mix,song2022self,mo2022localizing,mo2022closer,hu2020discriminative}.
Most previous approaches learn audio-visual correlations from third-person videos, while the distinct challenges of audio-visual learning in egocentric videos are underexplored.
%
%
Different from existing works, we propose an audio-visual learning framework to explicitly solve egomotion and out-of-view audio issues in egocentric videos.


%

\noindent \textbf{Egocentric video understanding.}
In the last decade, video scene understanding techniques thrived because of the well-defined third-person video datasets~\cite{carreira2017quo,soomro2012ucf101,caba2015activitynet,miech2019howto100m}.
Nevertheless, most of the algorithms are developed to tackle videos curated by human photographers.
The natural characteristics of egocentric video data, \eg, view changes, large motions,  and visual deformation, are not well-explored. 
To bridge this gap, multiple egocentric datasets~\cite{sigurdsson2018charades,damen2018scaling,su2016detecting,lee2012discovering,damen2020rescaling,grauman2021ego4d} have been collected. These datasets have significantly advanced investigations on egocentric video understanding problems, including activity recognition~\cite{kazakos2019epic,li2021ego,zhou2015temporal}, human(hand)-object interaction~\cite{cai2016understanding,nagarajan2019grounded,damen2016you,shan2020understanding}, anticipation~\cite{abu2018will,furnari2020rolling,liu2020forecasting,singh2016krishnacam}, and human body pose inferring~\cite{jiang2017seeing,ng2020you2me}.
However, only a handful of audio-visual works~\cite{cartas2019much,xiao2020audiovisual,kazakos2019epic, mittal2022learning} is presented for egocentric video understanding.
Among those, Kazakos~\etal~\cite{kazakos2019epic} proposed an audio-visual fusion network for action recognition, while Mittal~\etal~\cite{mittal2022learning} used audible interactions as cues to learn state-aware visual representations in egocentric videos.
There are limited studies in explicit egomotion mitigation and fine-grained audio-visual association learning in egocentric videos. 
%
Unlike past works, we tackle challenges in egocentric audio-visual data and propose a robust sounding object localization framework. To enable the research, we propose \textit{Epic Sounding Object} dataset based on Epic-Kitchens~\cite{damen2018scaling,damen2020rescaling}.
%


\section{Method}
\label{sec:method}

%
%
%

Our goal is to localize sounding objects in egocentric videos visually. We start by formulating our egocentric audio-visual object localization task in \cref{subsec:formulation}. 
Our proposed method includes a feature extraction process (described in \cref{subsec:extraction}), a two-stage cascaded feature enhancement pipeline (in \cref{subsec:cfe}), and a geometry-aware temporal aggregation module (explained in \cref{subsec:gatm}).
Finally, we summarize the overall training objective in \cref{subsec:learning}.

\subsection{Problem Formulation and Method Overview}
\label{subsec:formulation}
%
Given an egocentric video clip $V = \{I_i\}^T_{i=1}$ in $T$ frames and its corresponding sound stream $s$, sounding object visual localization aims at predicting location maps $\mathcal{O} = \{O_i\}^T_{i=1}$ that represent sounding objects in the egocentric video. 
Specifically, $O_i(x,y) \in \{0, 1\}$  and positive visual regions indicate locations of sounding objects.
In real-world scenarios, the captured sound can be a mixture of multiple sound sources $s = \sum_{n=1}^{N}s_n$, 
where $s_n$ is the $n$-th sound source and it could be out of view. 
For the visual input, the video frames may be captured from different viewpoints.
To design a robust and effective egocentric audio-visual sounding object localization system, we should consider the above issues in egocentric audio and visual data and answer two key questions: (Q1) how to associate visual content with audio representations while out-of-view sounds may exist;
(Q2) how to persistently associate  audio features with visual content that are captured under different viewpoints.

Due to the dynamic nature of egocentric videos, it is difficult and costly to annotate sounding objects for supervised training.
To bypass the tedious labeling, we solve the egocentric audio-visual object localization task in a self-supervised manner. The proposed framework is shown in Fig.~\ref{fig:framework}.
Our model first extracts representations from the audio $s$ and video clip $V$. 
In order to handle Q1, we develop a cascaded feature enhancement module to disentangle visually indicated sound sources 
and attend to visual regions that correspond to the visible sound sources.
To enable the disentanglement, we use on-screen sound separation task as the proxy and adopt a multi-task learning objective to train our model where the localization task is solved along with a sound-separation task. 
%
%
%
To deal with the egomotion in egocentric videos (Q2), we design a geometry-aware temporal modeling approach to mitigate the feature distortion brought by viewpoint changes and aggregate the visual features temporally.
%
We take the audio-visual temporal synchronization as the supervision signal and estimate the localization map $\Tilde{O_i}$.
%
%



%
\subsection{Feature Extraction}
\label{subsec:extraction}
%
\noindent
\textbf{Visual representation.} 
We use a visual encoder network $E_v$ to extract visual feature maps from each input frame $I_i$.
In our implementation, a pre-trained Dilated ResNet~\cite{yu2017dilated} model is adopted by removing the final fully-connected layer.
We can subsequently obtain a group of feature maps $v_i = E_v(I_i)$, where $v_i \in \mathbb{R}^{c \times h_v \times w_v}$.
Here $c$ is the number of channels, and $h_v \times w_v$ denotes the spatial size.

\noindent
\textbf{Audio representation.}
To extract audio representations from the input raw waveform, we first transform audio stream $s$ into a magnitude spectrogram $X$ with the short-time Fourier transform (STFT). 
Then, we extract audio features $a = E_a(X), a \in \mathbb{R}^{c \times h_a \times w_a}$ by means of a CNN encoder $E_a$ in the Time-Frequency (T-F) space. 
%
%

\subsection{Cascaded Feature Enhancement}
\label{subsec:cfe}

As discussed in Sec.~\ref{subsec:formulation}, a sound source $s_n$ in the mixture $s$ could be out of view due to constant view changes in egocentric videos and the limited FoV.
This poses challenges in visually localizing sound sources and performance can degrade when the audio-visual associations are not precise. 
%
%
%
%
%
%
To address this, we update the features in a cascaded fashion. We first force the network to learn disentangled audio representations from the mixture using visual guidance. Then we utilize the disentangled audio representations to inject the visual features with more localization cues.

\noindent
\textbf{Disentanglement through sound source separation.}
%
Sound source localization objective can implicitly guide the system to learn disentangled audio features as the network will try to precisely localize the sound, and in turn, the on-screen sound will get disentangled from the rest. 
%
%
However, we formulate our problem in an unsupervised setting where labels for such localization objective are not available.

Audio-visual sound separation task~\cite{gao2019co,zhao2018sound} uses visual information as guidance to learn to separate individual sounds from a mixture. Given the visual guidance, it is expected that the learned representations primarily encode information from visually indicated sound sources. Hence we argue for a multi-task learning approach to solve our primary task. Along with the audio-visual sounding object localization task, the network also learns to disentangle visible audio representations from the mixture through a source separation task. 

%

\begin{itemize}
    \vspace{-1mm}
    \item \textbf{Training.} We adopt the commonly used ``mix-and-separate" strategy~\cite{gao2019co,zhao2018sound} for audio-visual sound separation.
    Given the current audio $s^{(1)}$, we randomly sample another audio stream $s^{(2)}$ from a different video and mix them together to generate input audio mixture  $\tilde{s} = s^{(1)} + s^{(2)}$. 
    %
    We then obtain magnitude spectrograms $\tilde{X}$, $X^{(1)}$, $X^{(2)}$ for $\tilde{s}$, $s^{(1)}$ and $s^{(2)}$ respectively. 
    The audio features is then modified as $a = E_a(\tilde{X})$.

    \vspace{-1mm}
    \item \textbf{Inference.} During inference, we take the original audio stream as input: $s = s^{(1)}$ and $X = X^{(1)}$ to extract visually correlated audio representations. Note that the audio features is $a = E_a(X)$.
    %
    \vspace{-1mm}
\end{itemize}

We define the audio disentanglement network as a network $f(\cdot)$, which produces the disentangled audio features $\hat{a} \in \mathbb{R}^{c \times h_a \times w_a}$.
%
In this network, we want to associate the visual content with the audio representations to perform disentanglement in the embedding space.
%
%
Concretely, we first apply spatial average pooling on each $v_i$ and temporal max pooling along the time axis to obtain a visual feature vector $g_v \in \mathbb{R}^c$. 
Then we replicate the visual feature vector $h_a \times w_a$ times and tile them to match the size of $a$.
We concatenate the visual and audio feature maps along the channel dimension and feed them into the network.
Therefore, the audio feature disentanglement can be formulated as:
\begin{equation}
    \hat{a} = f(\,\textsc{Concat}[\, a, \; \textsc{Tile}(g_v)\,] \,).
\end{equation}
In practice, we implement the disentanglement network $f$ using two 1x1 convolution layers. The audio feature $\hat{a}$ will be used for both separation mask and sounding object localization map generation. 

%
%
To separate visible sounds, we add an audio decoder $D_a$ following the disentanglement network to output a binary mask $M_{pred} = D_a(\hat{a})$ (at the bottom of Fig.~\ref{fig:framework}).
U-Net architectures~\cite{gao2019co} are used in the audio encoder $E_a$ and decoder $D_a$. 
We implement the $E_a$ and $D_a$ in five convolution and up-convolution layers, respectively.
Details of the network architectures are provided in the Appendix.
The ground truth separation mask $M_{gt}$ can be calculated by determining whether the original input sound is dominant at locations $(u,v)$ in the T-F space: 
\begin{align}
    M_{gt}(u,v) = [X^{(1)}(u,v) \geq \Tilde{X}(u,v)].
\end{align}
%

To train the sound separator, we minimize the $\ell_2$ distance between the predicted and ground-truth masks as the disentanglement learning objective:
\begin{equation}
    \mathcal{L}_{dis} = ||M_{pred} - M_{gt}||_2^2 .
\end{equation}
%

\noindent
\textbf{Soft localization.}
%
%
Similar to the out-of-view sounds, the visual frames may contain sound-irrelevant regions.
In order to learn more precise audio-visual associations, we propose to highlight the spatial regions that are more likely to be correlated with the on-screen sounds by computing audio-visual attention.
The attention map will indicate the correlation between audio and visual representations at different spatial locations.
Given the output $\hat{a}$ from disentanglement network $f(\cdot)$, we apply max pooling on its time and frequency dimensions, obtaining an audio feature vector $g_{\hat{a}}$.
Then at each spatial position (x,y) of visual feature $v_i$, we compute the cosine similarity between audio and visual feature vectors:
\begin{equation}
    S_{i}: S_{i}(x,y) = \textsc{CosineSim}(v_i(x,y), g_{\hat{a}}).
\label{equ:avattn}
\end{equation}
\textsc{Softmax} is then used on $S_{i}$ to generate a soft mask that represents the audio-visual correspondence. 
Hence, each $v_i$ can be attended with the calculated weights:
\begin{align}
    \hat{v}_i = \textsc{Softmax}(S_{i}) \cdot v_i.
\end{align}


\subsection{Geometry-Aware Temporal Modeling}
\label{subsec:gatm}
\begin{figure}[t]
    \centering
    \includegraphics[width=0.45\textwidth]{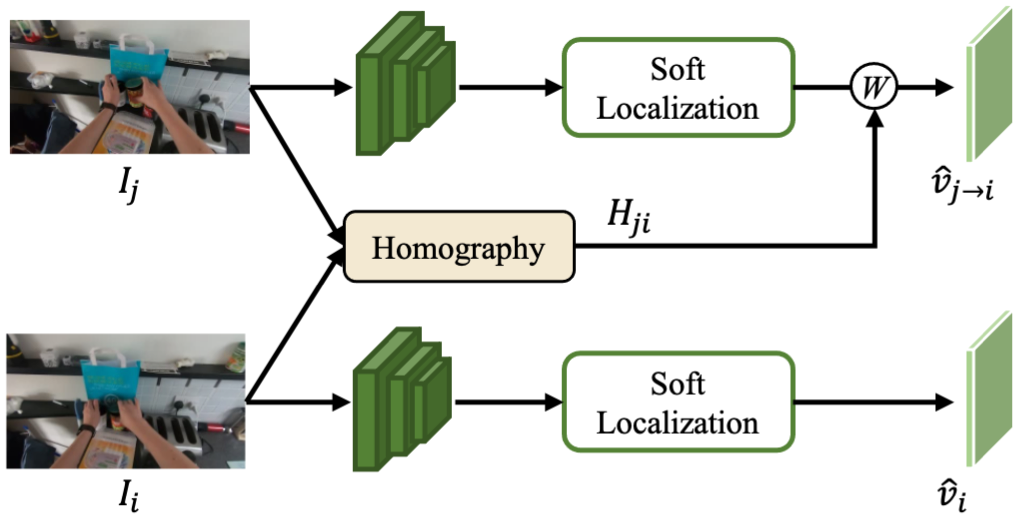}
    \caption{
    Overview of our proposed geometry-aware modeling approach. The visual features $\hat{v}_j$ are warped to viewpoint $i$ by homography transformation. 
    }
    \label{fig:gatm}
    \vspace{-4mm}
\end{figure}

%
Given the temporal nature of sounds and the persistence of audio-visual associations, we incorporate temporal information from neighboring frames to learn sounding object features.
%
%
However, temporal modeling is a challenging problem for egocentric videos due to widespread egomotion and object-appearance deformations.
%

%
Although visual objects are dynamically changing, the surrounding physical environment is persistent.
%
Hence, temporal variations in egocentric videos reveal rich 3D geometric cues that can recover the surrounding scene from changing viewpoints.  
Prior works have shown that given a sequence of frames, one can reconstruct the underlying 3D scene from the 2D observations~\cite{schonberger2016structure,vijayanarasimhan2017sfm}.
%
In our work, rather than reconstructing the 3D structures,  we estimate the relative geometric transformation between frames to alleviate egomotion. Specifically, we apply the transformation at the feature level to perform geometry-aware temporal aggregation.
Given $\{I_i\}^T_{i=1}$ and their features ${\{
\hat{v}_i\}^T_{i=1}}$, we take $\hat{v}_i$ as a query at a time and use the other features from neighboring frames as support features to aggregate temporal contexts.
For clarity, we decompose the geometry-aware temporal aggregation into two parts: geometry modeling and temporal aggregation.
%

\noindent
\textbf{Geometry modeling.}
This step aims to compute the geometric transformation that represents the egomotion between frames (see Fig.~\ref{fig:gatm}).
We found that homography estimation, which can align images taken from different perspectives, can serve as a way to measure geometric transformation.
We adopt SIFT~\cite{lowe2004distinctive} + RANSAC~\cite{fischler1981random} to solve homography.
To be specific, a homography is a $3\times3$ matrix that consists of 8 degree of freedom (DOF) for scale, translation, rotation, and perspective respectively.
Given the query frame $I_i$ and a supporting frame $I_j$, we use $h(\cdot)$ to denote the computation process:
\begin{equation}
    \mathcal{H}_{ji} = h(I_j, I_i)_{j\xrightarrow{} i} ,
\end{equation}
where $\mathcal{H}_{ji}$ represents the homography transformation from frame $I_j$ to $I_i$.
With the computed homography transformation, we can then apply it at the feature level to transform visual features $\hat{v}_j$ to $\hat{v}_{ji}$. The $\hat{v}_{ji}$ is egomotion-free under the viewpoint of $I_i$. 
Since the resolution of feature maps is scaled down compared to the raw frame size, the homography matrix $\mathcal{H}$ should also be downsampled using the same scaling factor.
The feature transformation can be written as:
\begin{equation}
     \hat{v}_{ji} = \mathcal{H}_{ji} \otimes \hat{v}_{j},
\end{equation}
where $\otimes$ represents the warping operation.
\begin{figure}[t]
    \centering
    \includegraphics[width=0.4\textwidth]{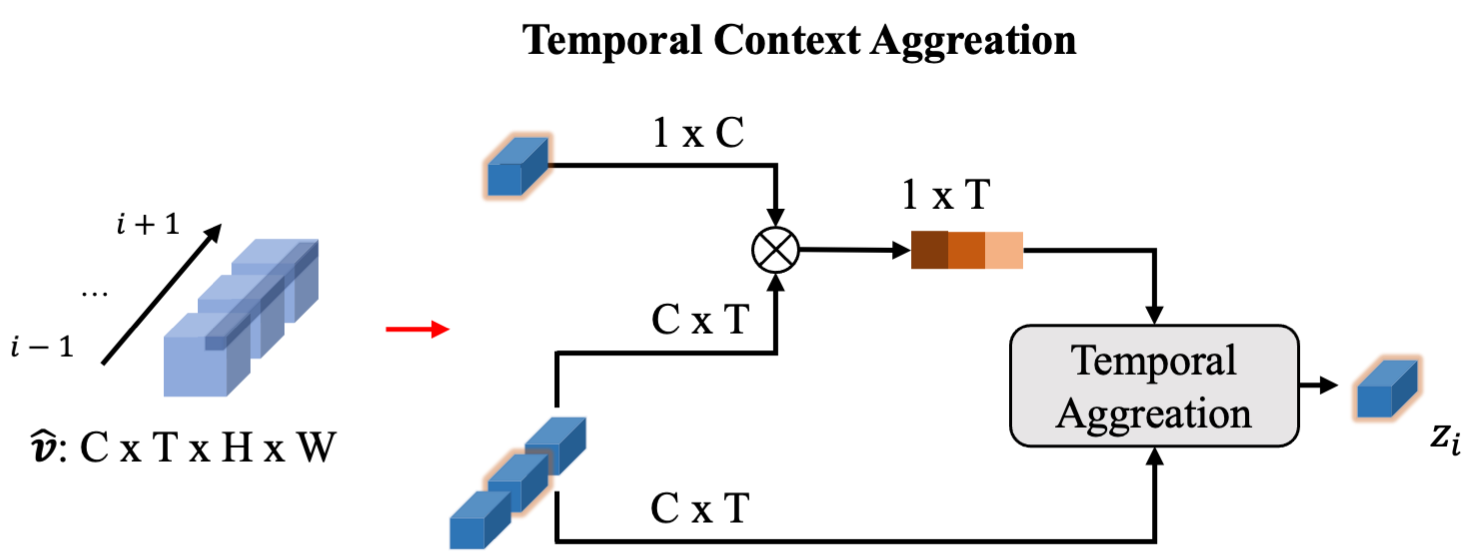}
    \caption{Illustration on the temporal context aggregation process for query feature $\hat{v}_{i}$ and neighboring frame features $\hat{v}_{i-1}$ and $\hat{v}_{i+1}$, which is performed independently at each spatial location. 
    }
    \label{fig:temporal}
    \vspace{-1em}
\end{figure}

\noindent
\textbf{Temporal aggregation.}
For the query feature $\hat{v}_{i}$, we end up with set of aligned features $\{\hat{v}_{ji}\}^{T}_{j=1}$ corresponding to each frame viewpoint. 
To aggregate the temporal contexts, we propose to compute the correlation between features from different frames at the same locations (see Fig.~\ref{fig:temporal}).
The aggregation process can be formulated as:
\begin{equation}
    z_i(x,y) = \hat{v}_i(x,y) + \textsc{Softmax}(\frac{\hat{v}_i(x,y) {\boldsymbol{\hat{v}}(x,y)}^T}{\sqrt{d}})\boldsymbol{\hat{v}}(x,y),
\end{equation}
where $\boldsymbol{\hat{v}} = [\hat{v}_{1i};...;\hat{v}_{Ti}]$ is the concatenation of frame features;
the scaling factor $d$ is equal to the feature dimension; and $(\cdot)^T$ represents the transpose operation~\cite{vaswani2017attention}.
The aggregation operation is applied at all spatial locations $(x,y)$ to generate the updated visual features $z_i$.

\begin{figure*}[!ht]
    \centering
    \includegraphics[width=0.98\textwidth]{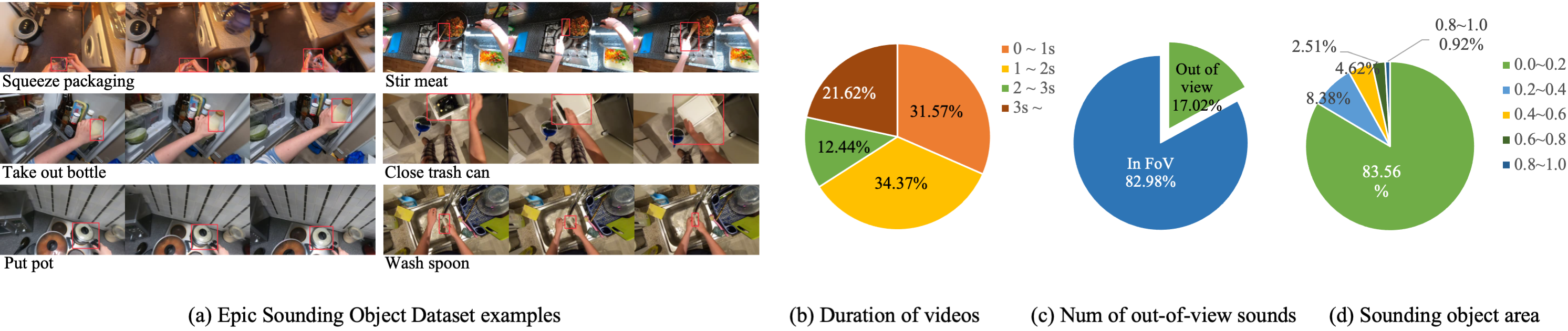}
    \vspace{-2mm}
    \caption{\textbf{Illustration of the \textit{Epic Sounding Object} dataset statistics}. \textbf{(a)} Example of our video frames and sounding object annotations. Class diversity (squeeze packaging, close trash can, put the pot, etc.) \textbf{(b)}: The distribution of untrimmed video duration.  \textbf{(c)}: The number of videos that are annotated as containing out-of-view sounds. \textbf{(d)}: Distribution of bounding box areas in Epic Sounding Object dataset, the majority of boxes cover less than 20\% of the image area, demonstrating the difficulty of this task. 
}
    \label{fig:dataset}
\end{figure*}

\subsection{Training Objective}
\label{subsec:learning}

We take audio-visual synchronization as the ``free" supervision and solve the task in a self-supervised manner using contrastive learning~\cite{afouras2020self,ots2018,senocak2018learning,chen2021localizing}.

With the audio feature vector $g_{\hat{a}}$ and the visual features $\{z_{i}\}^{T}_{i=1}$, we can compute an audio-visual attention map $S_i$ in Eq.~\ref{equ:avattn} for each frame $I_i$.
The training objective should optimize the network such that only the sounding regions have a high response in $S_i$. 
Since the ground-truth sounding map is unknown, we apply differential thresholding on $S_i$ to predict sounding objectness map ${O}_i = sigmoid((S_i - \epsilon)/\tau)$~\cite{chen2021localizing}, where $\epsilon$ is the threshold, and $\tau$ denotes the temperature that controls the sharpness. 

In an egocentric video clip, a visual scene is usually temporally dynamic. Sometimes a single audio-visual pair ($I_i$,$s$) may not be audio-visually correlated.
To this end, we solve the localization task in the Multiple-Instance Learning (MIL)~\cite{maron1997framework} setting to improve robustness.
Concretely, we use a soft MIL pooling function to aggregate the concatenated attention maps $\boldsymbol{S} = [S_1;...;S_T]$ by assigning different weights to $S_t$ at different time steps:
\begin{gather}
   \overline{S} = \sum_{t=1}^T (W_t \cdot \boldsymbol{S})[:,:,t],
\end{gather}
where $W_t[x,y,:] = \textsc{Softmax}(\boldsymbol{S}[x,y,:])$, $x$ and $y$ are the indices on spatial dimensions.
Subsequently, an aggregated sounding objectness map $\overline{O}$ is calculated from $\overline{S}$.
In this way, for each video clip $V$ in the batch, we can define its positive and negative training signals as:
\begin{gather}
    P = \frac {1}{|\overline{O}|} \langle \overline{O} , \overline{S}\rangle,
    \quad
    N = \frac {1}{hw} \langle \mathbf{1} , S_ {neg}\rangle,
\end{gather}
where $ \langle \cdot , \cdot \rangle$ is the Frobenius inner product. 

%
We obtain negative audio-visual attention maps $S_{neg}$ by associating the current visual inputs $I$ with audio from other video clips.
$\mathbf{1}$ denotes an all ones tensor with shape $h \times w$.
Therefore, the localization optimization objective is:
\begin{gather}
        \mathcal{L}_{loc} = -\frac{1}{N} \sum_{k=1}^N [\log \frac{\exp(P_k)}{\exp(P_k) + \exp(N_k)}],
\end{gather}
where $k$ is the video sample index in a training batch.
The overall objective is $\mathcal{L} = \mathcal{L}_{loc} + \lambda \mathcal{L}_{dis}$, where we empirically set $\lambda=5$ in our experiments.

\section{The Epic Sounding Object Dataset}
\label{sec:data}

Existing sound source visual localization evaluation datasets, such as SoundNet-Flickr~\cite{senocak2018learning}, VGG-Sound Source~\cite{chen2021localizing}, only contain \textit{third-person} recordings. To the best of our knowledge, there is no existing dataset that is suitable for evaluating our model.
%
%
Thus, we introduce an Epic Sounding Object Dataset for egocentric audio-visual sounding object localization. 
Built upon the well-known Epic-Kitchens~\cite{damen2020rescaling} dataset, we collect sounding object annotations on its action recognition test set.
%
%
\\
\textbf{Data preparation.} 
We select 13k test videos from the Epic-Kitchens action recognition benchmark as our source data. 
Since these videos are not originally collected for audio-visual analysis, they vary in length, and not all of them contain meaningful sounds.
%
%
%
To verify the videos for annotations, we conduct a two-step process:
We first determine if a video is silent by checking its sound-level in decibels relative to full scale. Consequently, silent videos are filtered out to provide a meaningful data source.
Second, we bin the videos by their duration and show the statistics in \cref{fig:dataset}~(b).
The majority last less than 2 seconds, and hence we choose to trim the center 1-second clip from each video.

After pre-processing, we obtain 5,089 videos in total for annotation.
For each video, we uniformly select three frames and annotate sounding objects in the frames. 
%
We follow previous works~\cite{tian2021cyclic,chen2021localizing} to use bounding boxes to annotate the objects that emit sounds.
%
%
%
%
To obtain proposals of potential sounding objects automatically, we follow Epic-Kitchens~\cite{damen2020rescaling} to use a Mask R-CNN object detector~\cite{he2017mask} trained on MS-COCO~\cite{lin2014microsoft} and a hand-objects detector~\cite{shan2020understanding} that is pretrained with 42K egocentric images~\cite{damen2020rescaling,li2015delving,sigurdsson2018charades}.
%
\begin{table}[t]
\centering
\begin{tabular}{p{1.5cm}l|ccc}
\toprule
\multicolumn{2}{c|}{Before voting}                 & \multicolumn{3}{c}{After voting}                                                \\ \midrule
Video                & \multicolumn{1}{c|}{Frames} & Video                & \multicolumn{1}{c}{Frames} & \multicolumn{1}{c}{Classes} \\
\multicolumn{1}{p{1.5cm}}{5,089} &      15,267                      & \multicolumn{1}{c}{3,172 } &      9,196                       &        30                  \\
\bottomrule
\end{tabular}
    \caption{Statistics of the \textit{Epic Sounding Object} dataset.}
    \label{tab:statistics}
    \vspace{-7mm}
\end{table}
\\
\textbf{Annotation collection.} 
Given the pre-processed data, we then annotate the sounding objects manually.
Unlike third-person view videos, the object-sound associations in the egocentric domain are more complicated.
There are two main challenges: 
(i) Numerous egocentric videos record wearer-environment interaction (\eg, a human places a dish on the table). The object-sound associations could be dynamic, and sometimes it is hard to determine what objects are emitting sounds;
and (ii) the objects in egocentric videos are often missing from the screen, resulting in variations in scale (see Fig.~\ref{fig:dataset}~(c)).
We address the above issues by taking advantage of human commonsense knowledge.
We ask three or more annotators from the Amazon Mechanic Turk to annotate the same video (frames).
Concretely, they do this by first watching the 1-second video with three annotated frames to confirm what objects make sounds in the video.
During the annotation course, they are asked to answer two questions:
 (1) Does the video contain out-of-view sounds?
 (2) Which bounding boxes correspond to the sounding objects?
%
We collect the annotations in multiple rounds until each video has at least three or more valid annotations from the Amazon annotators.
Finally, we conduct an annotation verification by voting on all videos. 
%
%
%
%
%
%
If at least two annotators agree on the same answer, it will be considered as a correct annotation;
If not, we will simply omit the video.
The annotation statistics after voting are shown in \cref{tab:statistics}. We obtain 30 classes of sounds by counting the noun (object) classes. The annotations are evenly split into two sets for validation and testing.
%
%
The examples and statistics in Fig.~\ref{fig:dataset} illustrate the diverse and complicated nature of egocentric audio-visual scenes.

\section{Experiment}

\begin{figure}[!t]
    \centering
    \includegraphics[width=0.47\textwidth]{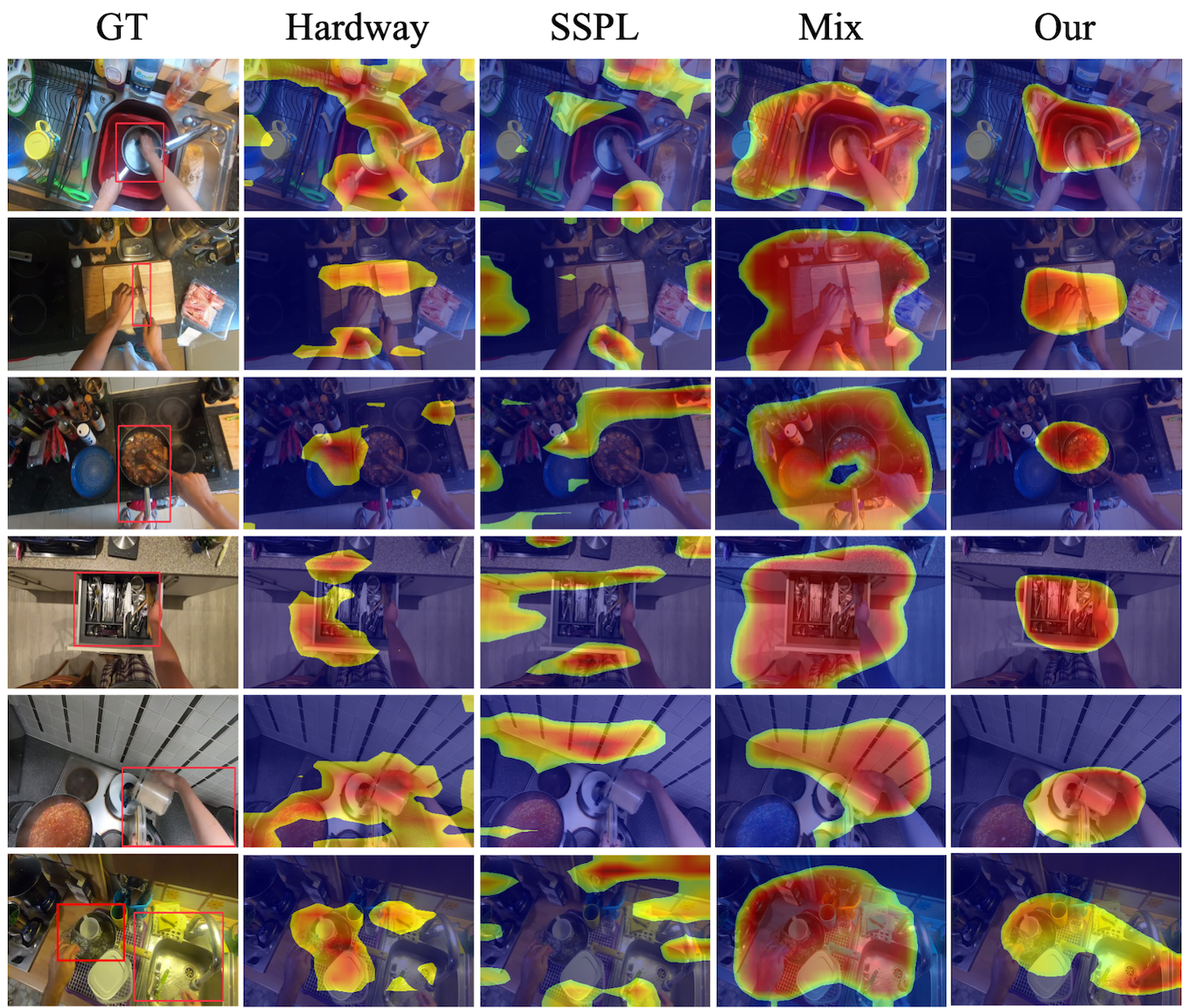}
    \vspace{-2mm}
    \caption{Qualitative comparison on \textit{Epic Sounding Object} dataset.
    We show diverse sounding objects in the kitchen scenes in the first column, sounding objects are annotated in red boxes.
    Our method outperforms all the compared works. 
    }
    \vspace{-4mm}
    \label{fig:viz}
\end{figure}
\noindent
\textbf{Datasets.} 
In our experiments, we use two egocentric datasets. 
%
%
(1) Epic-Kitchens~\cite{damen2020rescaling}: The dataset consists of 100 hours of egocentric recordings from 45 kitchen scenes. Thus, diverse kitchen-relevant events and sounds are in the dataset. We follow the same data split released in their action recognition benchmark and select 62,413 training videos.
We then filter out the silent videos by measuring the Decibels relative to full scale. 
This results in 47,214 training videos in total.
For evaluation, we use our annotated \textit{Epic Sounding Object} dataset to report the results;
(2) Ego4D~\cite{grauman2021ego4d}: Ego4D is the most recent large-scale egocentric video dataset.
Besides kitchen scenes, it includes diverse daily life scenarios.
Specifically, it consists of different subsets serving different benchmarks.
We select the ``Bristol" subset as it contains diverse scenarios (\eg, entertainment, sports, commuting, and more) to test our method.
We randomly sample and trim 50,000 1-second videos from this subset and use 90\%/10\% as the train/test split.
A similar filtering strategy is applied to obtain 26,858 videos for training.
We conduct experiments in Sec.~\ref{subsubsec:generalization} on this dataset to showcase the generalization ability of our method.
%
%
%
%
%
%
%
\\
\textbf{Evaluation metric.}
%
%
%
%
%
%
%
%
%
We follow the prior works~\cite{hu2022mix,chen2021localizing,li2021space} and adopt the pixel-level measurement for evaluating localization performance.  
Given the ground truth sounding object bounding boxes, we compute the Consensus Intersection over Union (CIoU) and Area Under Curve (AUC) between the predicted localization map and ground truth boxes. %
We report CIoU over a range of thresholds to expose the finer aspects of comparison.
%
\\
\textbf{Implementation details.}
To facilitate the training, we cut a 1-second long video around the center of each raw video.
We select the middle frame from the video clip and its four neighboring frames with an interval of 2 between frames.
Consequently, we get $T=5$ frames as visual input.
During training, the frames are first resized to 256$\times$256 and then randomly cropped to $224\times224$. 
During inference, all the frames are directly resized to the desired size without cropping.
For the audio stream, we extract the corresponding 1-second audio clip to create the audio-visual pairs.
The audio waveform is sub-sampled at 11kHz and transformed into a spectrogram with a Hann window of size 254 and a hop length of 64.
The obtained spectrogram is subsequently resampled to $128\times128$ to feed into the audio network. We set the number of audio and visual feature channels as 512 and choose $\epsilon=0.5$ and $\tau=0.03$. 
All models are trained with the Adam optimizer, with a learning rate of $10^{-4}$ on the visual encoder and temporal network, while using a learning rate of $10^{-2}$ for updating the audio encoder.

\begin{table}[t]
\centering
\scalebox{1.}{
\begin{tabular}{l||ccc|c}
\toprule
\multirow{2}{*}{}             & \multicolumn{3}{c|}{CIoU}                 &   \multicolumn{1}{c}{\multirow{2}{*}{AUC}}    \\ \cline{2-4}
                                               & @0.2  & @0.3  & \multicolumn{1}{c|}{@0.4} &    \\ \midrule
\multicolumn{1}{l||}{Attention~\cite{senocak2018learning}}                & 7.12      & -      &     -                      & 6.42     \\
\multicolumn{1}{l||}{STM~\cite{li2021space}}                                          & 12.10      & 7.64      & 4.01        & 8.87       \\
\multicolumn{1}{l||}{Hardway~\cite{chen2021localizing}}                                   & 24.51  & 13.55  & 6.10        & 13.38  \\
\multicolumn{1}{l||}{SSPL~\cite{song2022self}}                                            & 13.62      & 8.10      &  4.45            &  9.56     \\
\multicolumn{1}{l||}{Mix~\cite{hu2022mix}}                                         & 26.01     &  15.25    &  9.90              &  15.39     \\
\hline
\multicolumn{1}{l||}{Our}                                       &  \textbf{38.71}     &\textbf{ 19.42}      & \textbf{10.51  }                        & \textbf{18.38}     \\
\bottomrule
\end{tabular}
}
\vspace{-2mm}
\caption{Quantitative comparison of localization results on \textit{Epic Sounding Object} dataset. All methods are re-trained on Epic-Kitchen. The results of metrics CIoU@\{0.2, 0.3, 0.4\} and AUC are reported. The top-1 results are highlighted.}
\vspace{-4mm}
\label{tab:major}
\end{table}

\subsection{Results}
\label{subsec:results}

\subsubsection{Experimental Comparison}
\label{sec:expcomp}
To validate the effectiveness of our framework, we compare it with recent audio-visual localization methods: Attention~\cite{senocak2018learning}, STM~\cite{li2021space}, Hardway~\cite{chen2021localizing}, SSPL~\cite{song2022self} and Mix~\cite{hu2022mix}.
Among all the comparative methods, STM~\cite{li2021space} utilizes weak labels for the training, while the other methods are trained with self-supervision.
We hence adjust STM~\cite{li2021space} with our self-supervised localization loss.
As all the methods are developed for third-person view videos, we retrain their methods on our training data for a fair comparison.
The quantitative results are shown in Tab.~\ref{tab:major}.
We can find that our method outperforms all the compared approaches by a large margin in all metrics, indicating the benefits of mitigating out-of-view sounds and explicitly modeling egomotion in learning egocentric audio-visual localization.
Moreover, we provide a qualitative comparison to visually showcase our localization results.
In Fig.~\ref{fig:viz}, we can see that our model produces localization results that are tight around the ground truth sounding objects.

\begin{table}[!t]
\setlength{\tabcolsep}{1.1mm}
    \centering
    \begin{tabular}{cccccc||cc}
        \toprule
        \multicolumn{1}{c}{\multirow{2}{*}{Model}} & \multicolumn{3}{c}{TM} & \multicolumn{1}{c}{\multirow{2}{*}{SL}} & \multicolumn{1}{c||}{\multirow{2}{*}{$L_{dis}$}} & \multicolumn{1}{c}{\multirow{2}{*}{CIoU@0.2}} & \multicolumn{1}{c}{\multirow{2}{*}{AUC}}  \\ \cline{2-4}
        \multicolumn{1}{c}{} & Avg  & Max  & \multicolumn{1}{c}{GA} & & \multicolumn{1}{c||}{}   \\ 
        \midrule
         a & & & & &  & 27.41 & 15.36 \\
         b & \checkmark& &&&  & 31.84 & 15.79 \\
         c & &\checkmark &&&  & 33.29 & 16.10 \\
         d & & &\checkmark&&  & 37.38 & 16.59 \\
        f &&&  \checkmark&\checkmark &  & 38.21 & 17.92 \\
        g &&&  \checkmark&\checkmark & \checkmark   & \textbf{38.71} & \textbf{18.38} \\
        \bottomrule
    \end{tabular}
    \vspace{-2mm}
    \caption{Ablations on GATM, SL, and audio disentanglement module. The top-1 result in each column is highlighted. 
    }
    \vspace{-1em}
    \label{tab:ablation}
\end{table}

\vspace{-2mm}
\subsubsection{Ablation Study}
\label{sec:ablation}
\vspace{-2mm}
We conduct an ablation study to illustrate how each module affects localization performance. As shown in Tab.~\ref{tab:ablation}, we compare our full model with different baselines --- \textbf{model a}: we remove all the modules and only use the features from the visual and audio encoders to compute the localization map; \textbf{model b-d}: we insert \textit{Average}, \textit{Max}, and \textit{Geometry-Aware} Temporal modeling approaches separately in the framework; in \textbf{model f}, we incorporate the soft localization (SL) into the pipeline; and in \textbf{model g}, we employ the audio disentanglement module and train the model with $L_{dis}$.
%
By comparing \textbf{a} and \textbf{b-c}, we found that it's crucial to aggregate temporal context, while \textbf{d} emphasizes the importance of GATM in mitigating the egomotion in egocentric videos.
\textbf{model d} vs. \textbf{f} shows that SL slightly enhances the performance since some of the unrelated visual content can be reweighted. 
The comparison between \textbf{f} and \textbf{g} demonstrates that by incorporating audio feature disentanglement, the localization performance can be further boosted because it can handle the out-of-view sounds in videos. 

\noindent
\textbf{Naive baseline.} To assess the difficulty of the task, we provide a center box method that predicts a gaussian heatmap around the center. This results in a naive baseline of \textbf{16.51} compared to \textbf{27.41 (model a)} from \cref{tab:ablation}, showing that there are various challenging scenarios apart from the object being in the center that this naive baseline cannot capture.

\vspace{-2mm}
\subsubsection{Generalization to More Scenarios}
\begin{figure}[!hbpt]
    \centering
    \includegraphics[width=0.47\textwidth]{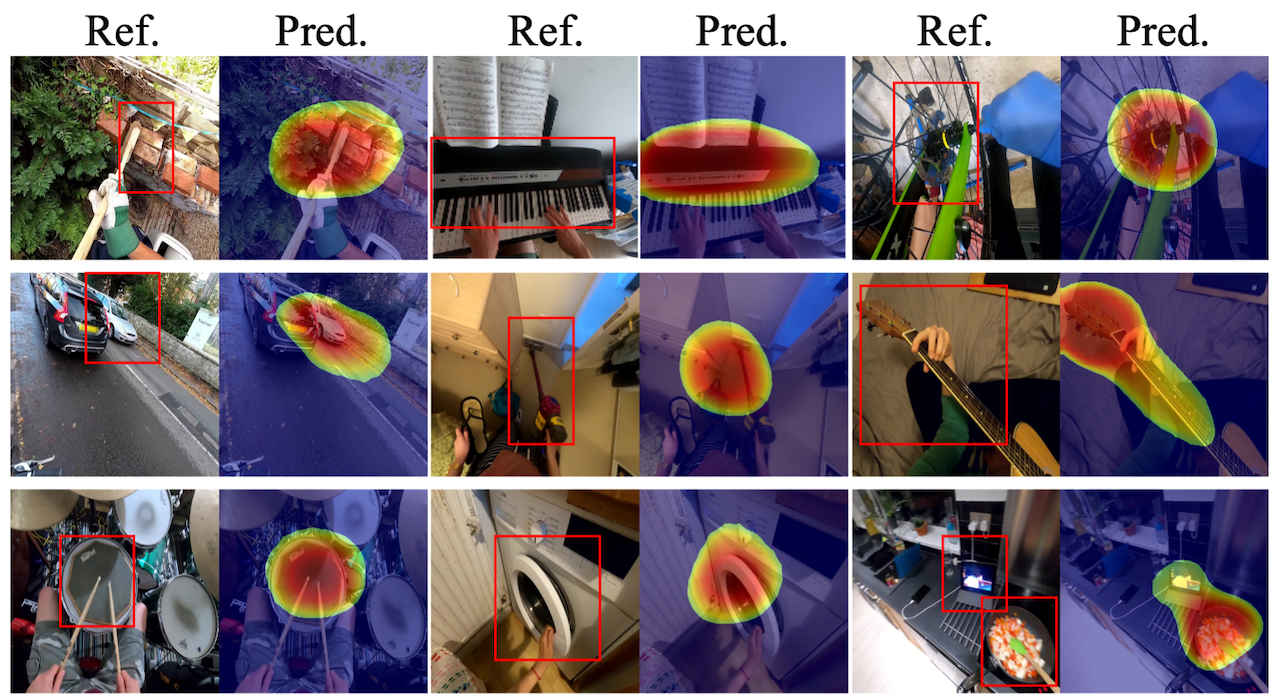}
    \caption{Localization results on diverse scenarios in Ego4D~\cite{grauman2021ego4d}.
    Ref.: Sounding objects; Pred.: predicted  localization results. 
    }
    \label{fig:ego4d}
\end{figure}

\vspace{-2mm}
\label{subsubsec:generalization}
The experiments on \textit{Epic Sounding Obejct} dataset demonstrate the effectiveness of our method in localizing sounding objects in the egocentric videos.
To further validate the generalization ability of our method, we train our audio-visual sounding object localization network on Ego4D~\cite{grauman2021ego4d} and qualitatively showcase the localization results in Fig.~\ref{fig:ego4d}.
The examples are all selected from the Ego4D test set.
We can see that our model can learn audio-visual associations and localize the sounding objects in diverse scenes.

\section{ Discussions and Conclusions}
\label{sec:conclusion}
In this work, we tackle a fundamental task: egocentric audio-visual localization to promote the field of study in egocentric audio-visual video understanding.
The uniqueness of egocentric videos, such as egomotions and out-of-view sounds pose significant challenges to learning fine-grained audio-visual associations.
To address these problems, we propose a new framework with a cascaded feature enhancement module to disentangle visually indicated audio representations and a geometry-aware temporal modeling module to mitigate egomotion.
%
%
Extensive experiments on our annotated \textit{Epic Sounding Object} dataset underpin the findings that explicitly mitigating out-of-view sounds and egomotion can boost localization performance and learn the better audio-visual association for egocentric videos.

\noindent
\textbf{Limitations.} The proposed geometry-aware temporal modeling approach requires geometric transformation computation. 
For certain visual scenes with severe illumination changes or drastic motions, the homography estimation may fail. Then, our GATM will degrade to a vanilla temporal modeling approach. To mitigate the issue, we can consider designing a more robust geometric estimation approach.

\noindent
\textbf{Potential Applications.} Our work offers potential for several applications: (a) Audio-visual episodic memory. As egocentric video records what and where of an individual’s daily life experience, it would be interesting to build an intelligent AR assistant to localize the object (``\textit{where did I use it?}'') by processing an audio query, \eg, an audio clip of ``vacuum cleaner''; (b) Audio-visual object state recognition. In egocentric research, it is important to know the state of objects that human is interacting with, while the human-object interaction often makes a sound. Therefore, localizing objects by sounds provides a new angle in recognizing an object state; (c) Audio-visual future anticipation: following the audio-visual object state recognition task, it's natural to predict the trajectory of a sounding object by analyzing the most recent audio-visual clips.


{\small
\bibliographystyle{ieee_fullname}
\bibliography{egbib}
}

\clearpage

\setcounter{figure}{7}
\setcounter{table}{3}



\thispagestyle{empty}
\appendix


\section*{Appendix}

\section{Video Demo}
In our video demo, we illustrate our motivation and method. Next, we show (a) localization results on the Epic Sounding Object dataset, and (b) on the Ego4D dataset.


\section{Details of Our Framework}
Our framework consists of a visual branch and an audio branch. We start with elaborating on the audio branch and then move to the visual one.

\begin{figure}[!h]
    \centering
    \includegraphics[width=0.49\textwidth]{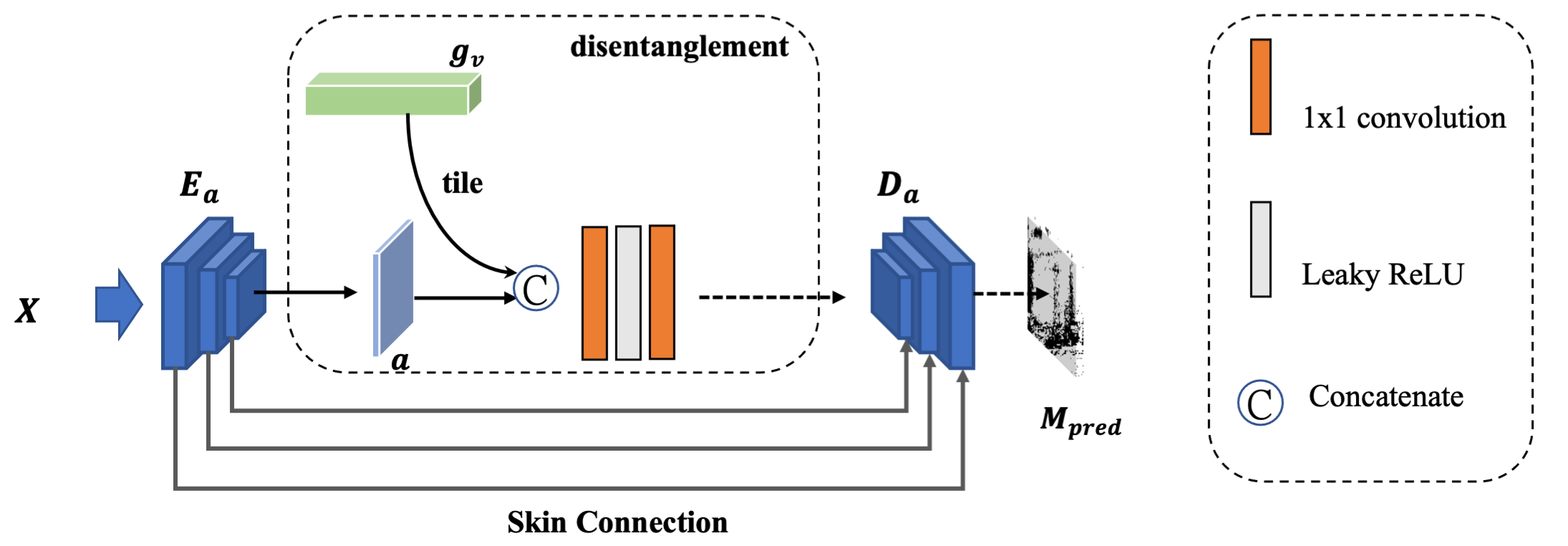}
    \caption{Details of audio branch, including an encoder $E_a$, a disentanglement network $f$ and a decoder $D_a$.}
    \label{fig:disentanglement}
\end{figure}

\noindent
\textbf{Audio Encoder $E_a$ and Decoder $D_a$.} To extract the audio feature from input spectrogram $X$ and predict the separation mask, our audio branch contains an encoder $E_a$ and a decoder $D_a$ as shown in Fig.~\ref{fig:disentanglement}. In practice, we design the encoder and decoder in a U-NET style architecture, i.e., skin connections are enabled. The encoder-decoder network consists of five convolution layers and five up-convolution layers, and all layers adopt a 4x4 kernel with stride 2. A BatchNorm~(BN) layer and a ReLU activation layer are appended after each convolution/up-convolution layer. For the last layer in $D_a$, we use a Sigmoid layer instead of the ReLU and remove the BN layer to output the mask. Skip connections allow the information flow from layer $i$ in $E_a$ to layer $n-i$ in $D_a$, where $n=5$ is the total number of layers. 

\noindent
\textbf{Disentanglement Network.}
The disentanglement network consists of two 1x1 convolution layers to obtain the visually indicated audio features. Given the audio input features $a$ with a size of 4x4x512, we first tile the visual feature $g_v \in \mathbb{R}^{512}$ by 4x4 times along the Time and Frequency axes to match the size of $a$. Then we concatenate audio and visual features along the channel dimension. The concatenated features will go through two 1x1 convolution layers with Leaky ReLU in between. The output feature maps are also of size 4x4x512.

In the following, we will explain some details about our visual branch.

\noindent
\textbf{Visual Encoder $E_v$.}
The visual branch includes a visual encoder $E_v$ to extract the features in the beginning. It takes $T=5$ frames of dimension 224x224x3 as inputs and outputs 5 feature maps of dimension 28x28x512. We use a pre-trained Dilated ResNet to implement $E_v$.

\noindent
\textbf{Geometry-Aware Temporal Modeling Module.}
Given frames $I_i$ and $I_j$ and their corresponding features $v_i$ and $v_j$, we first estimate the homography transformation $\mathcal{H}_{ji}$ between frames. The homography is a 3x3 matrix: 

\begin{equation}
\mathcal{H}_{ji} = 
    \begin{bmatrix}
    h_{11} & h_{12} & h_{13}\\
    h_{21} & h_{22} & h_{23} \\
    h_{31} & h_{32} & h_{33}
\end{bmatrix}
\end{equation}

\noindent
$I_j$ can thus be warped to $I_{ji}$ as $I_{ji} = \mathcal{H}_{ji} \otimes {I}_{j}$. Instead of applying homography at the image level, we use it in the feature space to warp the features. If the homography estimation fails due to the poor image condition or drastic viewpoint change, $\mathcal{H}_{ji}$ will be replaced with an identity matrix:
\begin{equation}
\mathcal{H}_{ji} = 
    \begin{bmatrix}
    1 & 0 & 0\\
    0 & 1 & 0 \\
    0 & 0 & 1
\end{bmatrix}
\end{equation}

\noindent
Therefore, our GATM will degrade to a vanilla temporal modeling approach but still can leverage the temporal contexts.
The temporal context aggregation is implemented as single-head attention at the time dimension. 

\section{Analysis on Audio Disentanglement}

\begin{figure}[ht]
    \centering
    \includegraphics[width=0.47\textwidth]{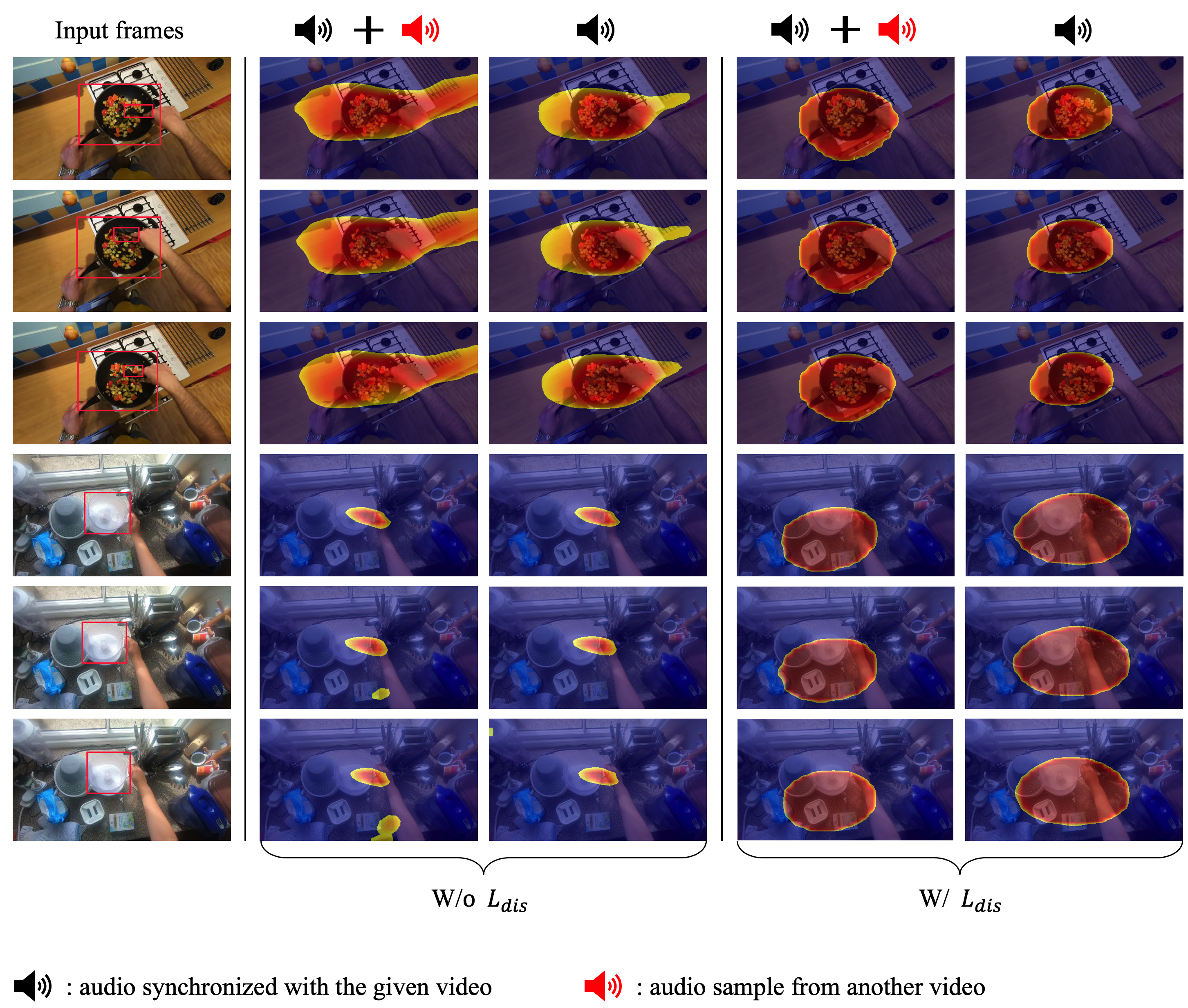}
    \caption{Visualizations on localization results generated from the model without (column 2-3) and with (column 4-5) audio disentanglement. Our model trained with $L_{dis}$ can yield consistent results when dealing with and without visually irrelevant audio in the input.}
    \label{fig:oov}
\end{figure}

\begin{table}[ht]
\setlength{\tabcolsep}{2mm}
    \centering
    \begin{tabular}{c|c|ccc}
        \toprule
        Model & Input  & @0.2 & @0.3 & @0.4   \\
        \midrule
        w/o $L_{dis}$ & $s^{(1)}+s^{(2)}$ & 38.15 & 18.60 & 10.42 \\
        w/ $L_{dis}$ & $s^{(1)}+s^{(2)}$ & 38.70 & 19.42 & 10.51 \\
        w/o $L_{dis}$ & $s^{(1)}$ & 38.21 & 18.67 & 10.42 \\
        w/ $L_{dis}$ & $s^{(1)}$ & 38.71 & 19.42 & 10.51 \\
        \bottomrule
    \end{tabular}
    \caption{Analysis of our model's performance by controlling the out-of-view sounds. We compare two models (trained w/ and w/o $L_{dis}$) and report results over CIoU@\{0.2, 0.3, 0.4\}. 
    }
    \label{tab:noise}
\end{table}

\begin{figure*}[ht]
    \centering
    \includegraphics[width=0.9\textwidth]{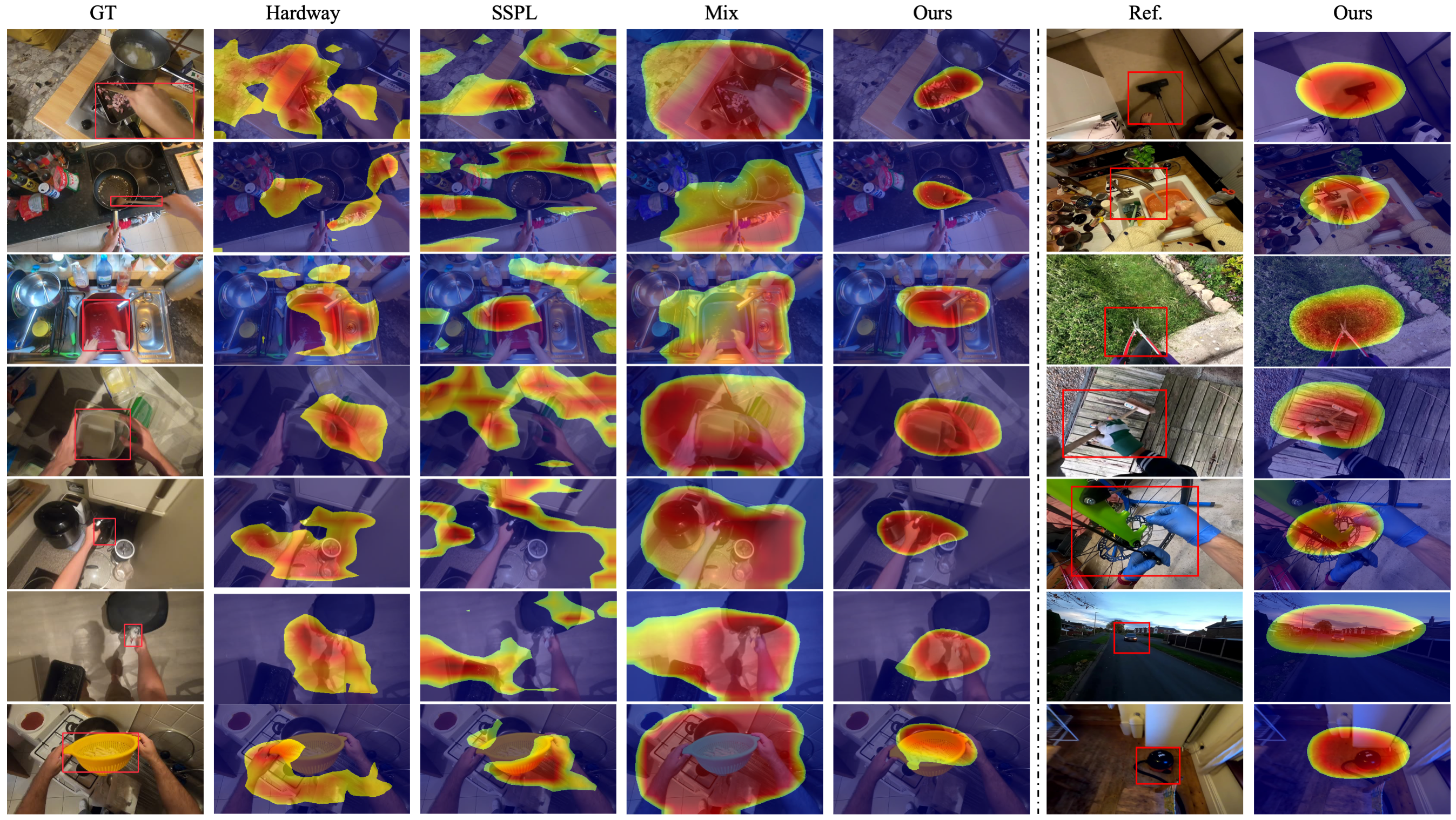}
    \caption{More visualization results of comparative methods and our approach on both Epic Sounding Object dataset (column 1-5) and Ego4D~\cite{grauman2021ego4d} dataset (column 6-7). The corresponding sounding objects are shown in red boxes (columns 1 and 6). Our method can produce more precise localization results and generalize to diverse daily scenarios.}
    \label{fig:more}
\end{figure*}

To further investigate the issue of out-of-view sounds in egocentric videos, we experimented with analyzing the performance of our model when dealing with audio mixtures.
We compare two different models. One is trained with disentanglement loss $\mathcal{L}_{ids}$ while the other does not.
During inference, we generate two different audio inputs: one is by mixing the original audio $s^{(1)}$ with an audio clip $s^{(2)}$ sampled from another video in the dataset, while the other one takes the original audio $s^{(1)}$ as input.
The corresponding results are shown in Table~\ref{tab:noise}.
We found that with audio feature disentanglement, our model achieves stable performance for both mixed and original audio inputs.
Visualizations in Fig.~\ref{fig:oov} also demonstrate the effectiveness of disentanglement training.


\section{More Visualizations}
We visualize more localization results for examples in the Epic Sounding Object and Ego4D~\cite{grauman2021ego4d} datasets (shown in Fig.~\ref{fig:more}). The figure shows that our framework can correctly localize various sounding objects, e.g., pan, spoon, box, vacuum cleaner, scissor, car and etc. 
Both indoor and outdoor scenarios are covered.  An example of people watching independent activity is also shown (Row 6 for Ego4D results).

\section{More Experiments}
In this section, we provide extra ablations and hyper-parameter experiments to add to the thoroughness of the evaluation. 

\begin{table}[ht]
\setlength{\tabcolsep}{2mm}
    \centering
    \begin{tabular}{ccc}
        \toprule
        Baseline & Baseline + SL  & Baseline + $L_{dis}$   \\
        \midrule
        27.41 & 28.83 & 32.96 \\
        \bottomrule
    \end{tabular}
    \caption{Analysis of the efficacy of Soft Localization (SL) and disentanglement loss $L_{dis}$. We report results on CIoU@0.2 metric. 
    }
    \label{tab:more_ablation}
\end{table}

\noindent
\textbf{Detailed ablations.} We present a more detailed ablation to complement Table 3 in the main paper. Specifically, we add Soft Localization or $L_{dis}$ to the baseline model before adding the geometry-aware temporal modeling module. Results are reported in Table~\ref{tab:more_ablation}. The SL module can slightly increase the performance to 28.83 (5.2\%$\uparrow$). When the disentanglement module is added to the baseline model without GATM, the performance is boosted to 32.96 (20.2\%$\uparrow$), which validates the effectiveness of audio disentanglement for solving out-of-view sounds problem.

\begin{table}[ht]
\setlength{\tabcolsep}{2mm}
    \centering
    \begin{tabular}{ccc}
        \toprule
        $\lambda=1$ & $\lambda=5$  & $\lambda=10$   \\
        \midrule
        30.85 & 32.96 & 32.10 \\
        \bottomrule
    \end{tabular}
    \caption{Hyper-parameter evaluation on the coefficient that controls the impact of disentanglement loss $L_{dis}$. The results are reported on model ``\textit{Baseline + SL}''.
    }
    \label{tab:lambda}
\end{table}

\begin{figure*}[!ht]
    \centering
    \includegraphics[width=0.85\textwidth]{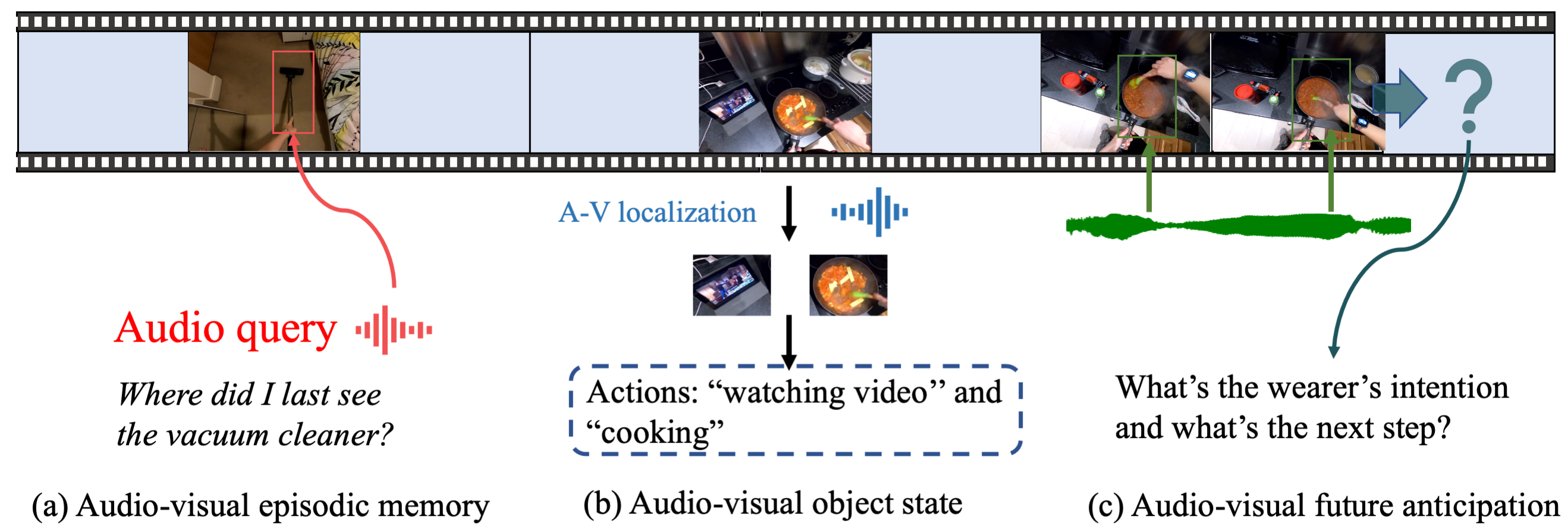}
    \caption{An overview of potential applications following egocentric audio-visual object localization, including (a) audio-visual episodic memory, (b) audio-visual object state, and (c) audio-visual future anticipation. These applications target understanding the \textit{past}, \textit{current}, and \textit{future} of the wearer's experience. Such capability is enabled by taking fine-grained audio-visual perception.}
    \label{fig:discussion}
\end{figure*}

\noindent
\textbf{Hyper-parameter $\lambda$.} In Table~\ref{tab:lambda}, we evaluate the effect of $\lambda$ that is used to balance the losses. When $\lambda$ is small, the model is not sufficiently trained to remove the visually unrelated contents from the audio representation, yielding inferior results. However, as $\lambda$ becomes larger ($\lambda=10$), the training objective will focus more on separation instead of accurate localization. Therefore, we select $\lambda=5$ empirically for our main model.

\begin{table}[ht]
\setlength{\tabcolsep}{2mm}
    \centering
    \begin{tabular}{ccc}
        \toprule
        $T=3$ & $T=5$  & $T=7$   \\
        \midrule
        36.46 & 37.38 & 35.34 \\
        \bottomrule
    \end{tabular}
    \caption{Hyper-parameter evaluation on the number of frames $T$.  The results are reported on model ``\textit{Baseline + GATM}''.
    }
    \label{tab:num_frame}
\end{table}

\noindent
\textbf{Hyper-parameter $T$.} The number of frames $T$ used to aggregate the temporal context is important and could further demonstrate the usefulness of the geometry-aware temporal modeling module. In Table~\ref{tab:num_frame}, we show quantitative results with various frame numbers. We can find that aggregating temporal information is significant as the performance boost from $T=3$ to $T=5$. When the frame number becomes even larger, the difficulty in effectively aligning visual frame features increases as greater viewpoint changes happen. Therefore, we choose $T=5$ in our main experiments.

\begin{figure*}[!htbp]
    \centering
    \includegraphics[width=0.95\textwidth]{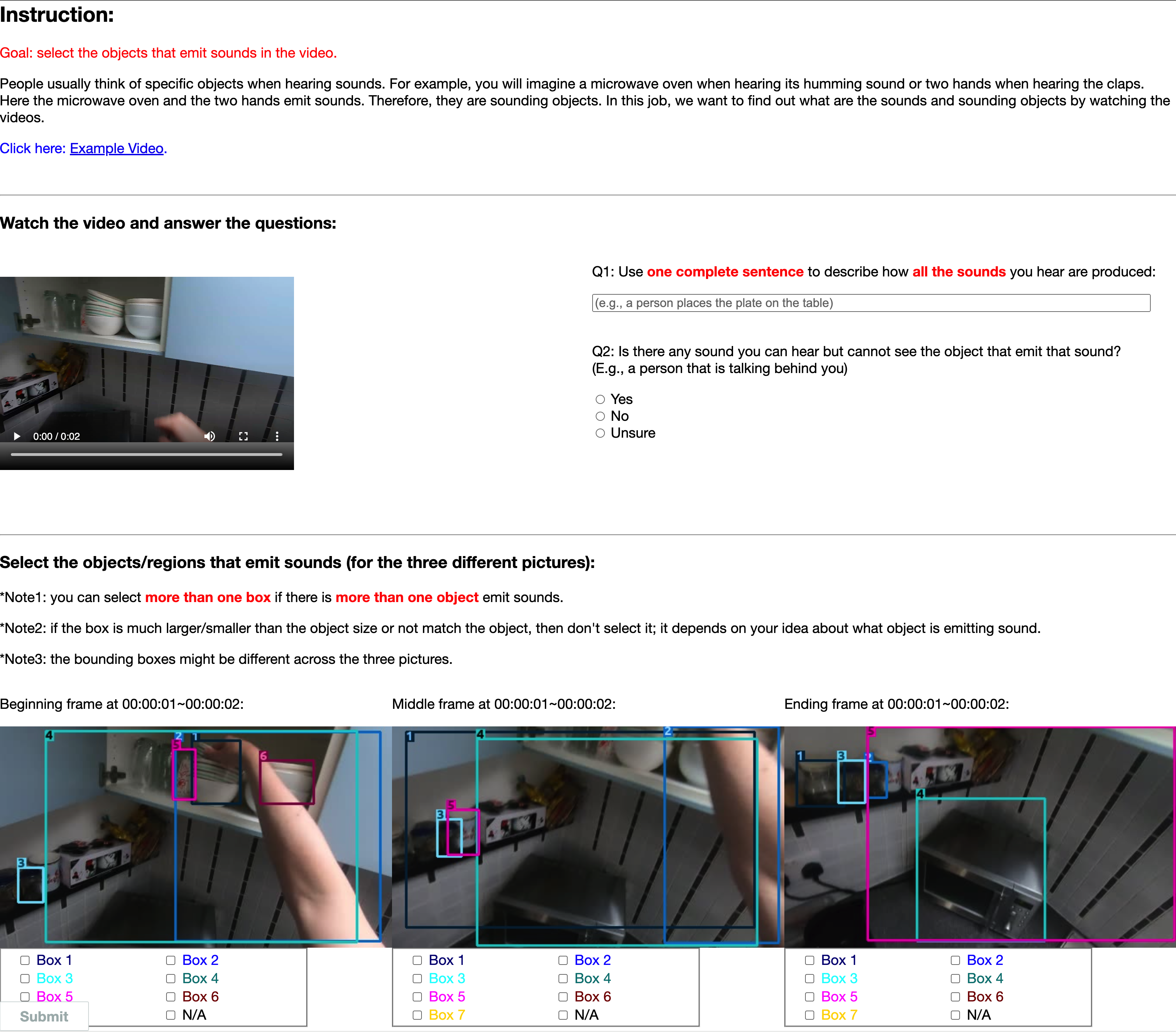}
    \caption{Example of our annotation interface. We ask every Amazon Mechanic Turk worker to watch the video first. Then, they are required to answer two questions and annotate the sounding objects correspondingly. The rules of selecting sounding objects ensure the quality of annotations. Besides, we conduct a voting process to obtain precise annotations for each video. } 
    \label{fig:collection}
\end{figure*}

\section{Potential Applications}

Our work has the potential to facilitate a range of applications, which are described below:

\noindent
\textbf{Audio-Visual Episodic Memory.}
As egocentric video records what and where of an individual’s daily life experience, it would be interesting to build an intelligent AR assistant to search the object that has been presented in the past. Previously, the episodic memory task could take an image or language query as input to localize the object. The correlation between visual objects and sound is less explored. Our egocentric audio-visual object localization task can extend episodic memory with auditory sense. As shown in Fig.~\ref{fig:discussion}~(a),  people can ask ``where did I use the vacuum cleaner?'', and the AR assistant can give the answer by feeding an audio query (a vacuum cleaner audio clip from the web) to the localization network. Therefore, the vacuum cleaner can be found in the video.

\noindent
\textbf{Audio-Visual Object State.}
In egocentric research, it is important to know the state of objects that humans interact with. The recognized object's state helps understand the wearer's actions. 
Interestingly, human-object interaction often makes a sound. Therefore, localizing objects by sounds provides a new angle in recognizing an object's state. For example, in Fig.~\ref{fig:discussion}~(b), both the laptop and the pot make sounds. By feeding the audio and the frame to the localization model, the laptop and the pot are localized correspondingly, which indicates that both objects are in a ``working'' state. Consequently, the ``watch video'' and ``cooking food using a pot'' actions are easily detected.

\noindent
\textbf{Audio-Visual Future Anticipation.}
As the audio-visual object state task indicate the human activity at the ``current'' moment by utilizing the sounding object results, it is natural to predict the future by analyzing the most recent audio-visual clips. The sound may change continuously as the object's state changes. In Fig.~\ref{fig:discussion}~(c), when the cooking is completed, the frying pot sound will be different and hence indicate the wearer's future action: the wearer may move the pot to the table or pour food into a bowl. Thus, by analyzing the audio-visual object state changes, the future of the wearer can be anticipated.

\section{Epic Sounding Object Dataset}

\noindent
\textbf{Amazon Mechanic Turk Annotation Collection.}
Annotating sounding objects in egocentric videos is challenging. Sounds are often correlated with human-object interactions, and sounding objects are sometimes occluded or under severe deformation due to frequent viewpoint changes. Therefore, annotating sounding objects automatically is difficult. To this end, we follow a semi-automatic labeling process by first generating bounding boxes for potential sounding objects. We use a Mask R-CNN object detector~\cite{he2017mask} trained on MS-COCO~\cite{lin2014microsoft} and a hand-objects detector~\cite{shan2020understanding} that pretrained with 42K egocentric images~\cite{damen2020rescaling,li2015delving,sigurdsson2018charades} to produce bounding boxes. Second, we manually annotate the sounding objects. Due to the above-described difficulty, people may have different opinions about what objects are emitting sounds. To reduce this uncertainty, we ask three or more people to annotate the same video and apply a voting process to the annotations. 

Specifically, we take advantage of the Amazon Mechanic Turk to label the sounding objects. We develop an interface (as shown in Fig.~\ref{fig:collection}) to support this process. First, the annotator is required to watch the video to ensure that sounding objects correspond to the sounds. Then the annotator will answer the following questions: (1) \textit{Use one complete sentence to describe how all the sounds you hear are produced.} Sound sometimes is ambiguous to annotate in bounding boxes when it is made by the interaction between objects, e.g., putting down the dish on the table. Whether ``only dish'' or  ``both dish and table'' are the sounding object is hard to determine. In this case, a language description is suitable to handle the ambiguity; (2) \textit{Whether the video contains sound you can hear but cannot see.} The answer hints at whether out-of-view sounds exist in the video. In egocentric videos, out-of-view sounds might be created due to the limited FoV. Therefore, we include this question to provide statistical analysis about the out-of-view sound problem; (3) by watching the video, the annotators are required to select the bounding boxes that correspond to the objects that emit sounds. The goal is to select the bounding box that humans recognize as a sounding object. All the collected answers will be passed to a voting process to determine the final annotations for each video (see Fig.~5 in the main paper).



\end{document}